\documentclass{article}
\usepackage[export]{adjustbox}
\usepackage{amsmath}
\usepackage{booktabs}
\usepackage{booktabs}
\usepackage{multirow}
\usepackage{graphicx}
\usepackage{wrapfig}
\usepackage{algorithm}
\usepackage{algpseudocode}
\usepackage{natbib}
\usepackage{makecell}
\usepackage{subcaption}
\usepackage{caption}      
\usepackage{booktabs}     
\usepackage{graphicx}     
\usepackage{array}       
\usepackage{booktabs}
\usepackage{array}
\usepackage{amsmath}
\usepackage{amssymb}
\usepackage{amsfonts}
\usepackage{multirow}
\usepackage{verbatim}
\usepackage{caption}
\usepackage{longtable}
\usepackage{supertabular}
\usepackage{CJKutf8}
\usepackage[utf8]{inputenc} 
\usepackage{adjustbox}
\usepackage[T1]{fontenc}
\usepackage[french,vietnamese,mongolian,greek,english]{babel}
\usepackage{pifont}
\usepackage[table]{xcolor}
\usepackage{enumitem}
\usepackage{tablefootnote}
\usepackage{xspace}
\usepackage{textcomp}
\usepackage{makecell}
\usepackage{lscape} 
\usepackage{siunitx}
\usepackage{listings}
\usepackage{xcolor}
\usepackage{wrapfig}
\usepackage{iclr2025_conference,times}

\usepackage{amsmath,amsfonts,bm}

\def\eqref#1{equation~\ref{#1}}

\def\1{\bm{1}}

\DeclareMathAlphabet{\mathsfit}{\encodingdefault}{\sfdefault}{m}{sl}
\SetMathAlphabet{\mathsfit}{bold}{\encodingdefault}{\sfdefault}{bx}{n}

\usepackage{tikz}
\usepackage[most]{tcolorbox}
\usepackage{graphicx}
\usepackage{caption}

\usetikzlibrary{positioning}

\definecolor{gblue9}{rgb}{0.2, 0.5, 0.7}
\usepackage{hyperref}
\usepackage{url}
\usepackage{xspace}

\usepackage{calligra}
\usepackage{fontenc}

\newcommand{\eg}{e.g.,\xspace}
\newcommand{\ie}{i.e.,\xspace}

\newcommand{\xvl}{R-4B}
\newcommand{\xvlbase}{R-4B-Base}
\newcommand{\xvlrl}{R-4B-RL}

\usepackage{graphicx} 
\vspace{-0.5cm}
\title{\includegraphics[height=1.5em, valign=c]{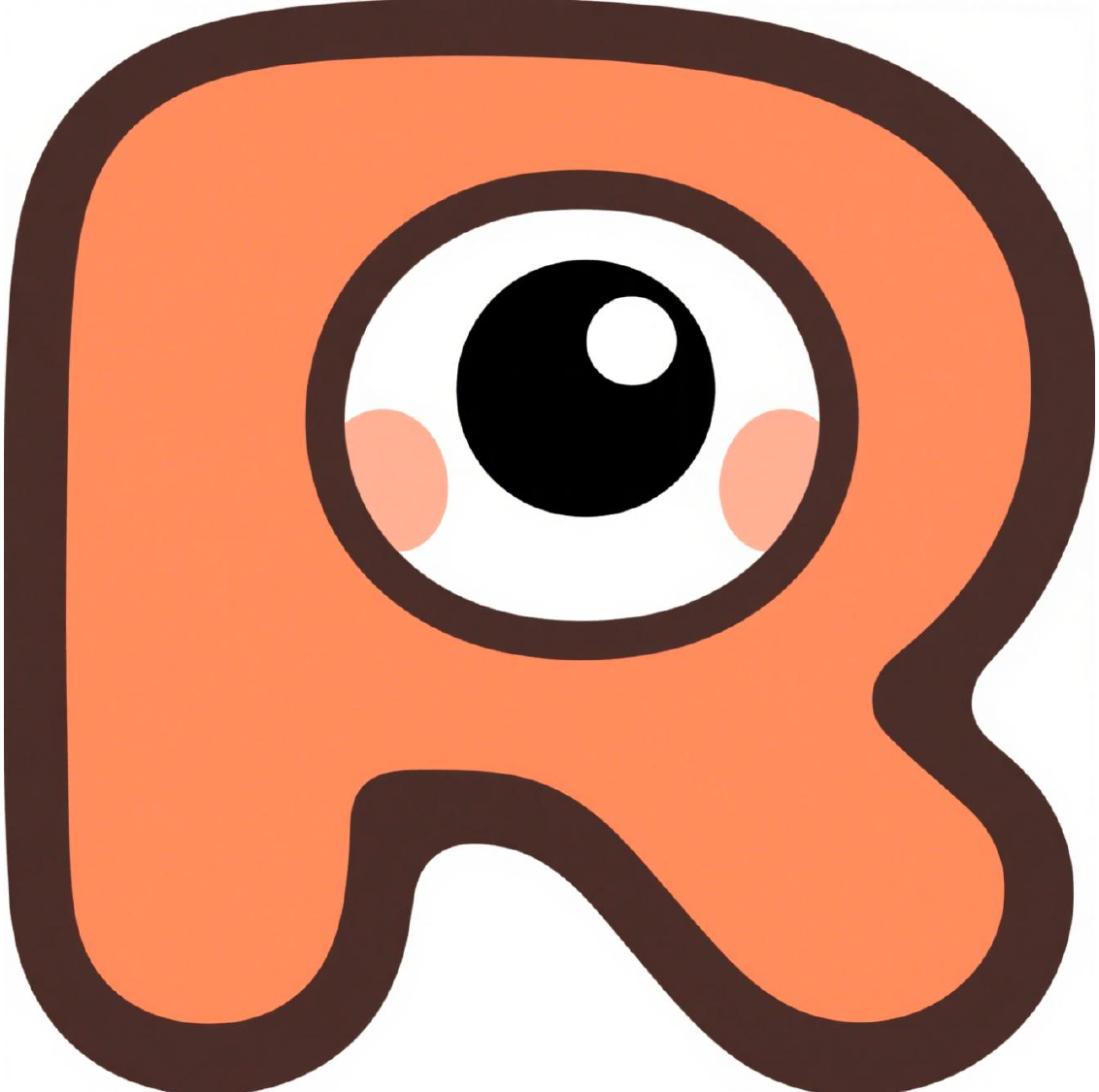}-4B: Incentivizing General-Purpose Auto-Thinking Capability in MLLMs via Bi-Mode Annealing and Reinforce Learning}

\author{\vspace{-0.8cm}\\
{\large\textsuperscript{1}Tencent Hunyuan Team \qquad \textsuperscript{2}Institute of Automation, CAS} \\
\vspace{-0.3cm} \\
\github ~ \url{\ghlink} \\  \huggingface ~ \url{\hflink}
}

\iclrfinalcopy 
\begin{document}

\maketitle

\begin{abstract}

Multimodal Large Language Models (MLLMs) equipped with step-by-step thinking capabilities have demonstrated remarkable performance on complex reasoning problems. 
However, this thinking process is redundant for simple problems solvable without complex reasoning.
To address this inefficiency, we propose \xvl{}, an auto-thinking MLLM, which can adaptively decide when to think based on problem complexity.
The central idea of \xvl{} is to empower the model with both thinking and non-thinking capabilities using bi-mode annealing, and apply Bi-mode Policy Optimization~(BPO) to improve the model's accuracy in determining whether to activate the thinking process.
Specifically, we first train the model on a carefully curated dataset spanning various topics, which contains samples from both thinking and non-thinking modes.
Then it undergoes a second phase of training under an improved GRPO framework, where the policy model is forced to generate responses from both modes for each input query.
Experimental results show that \xvl{} achieves state-of-the-art performance across 25 challenging benchmarks. It outperforms Qwen2.5-VL-7B in most tasks and achieves performance comparable to larger models such as Kimi-VL-A3B-Thinking-2506 (16B) on reasoning-intensive benchmarks with lower computational cost.

\end{abstract}

\vspace{-0.3cm}

\begin{figure*}[h!]
\centering
\includegraphics[width=1\linewidth]{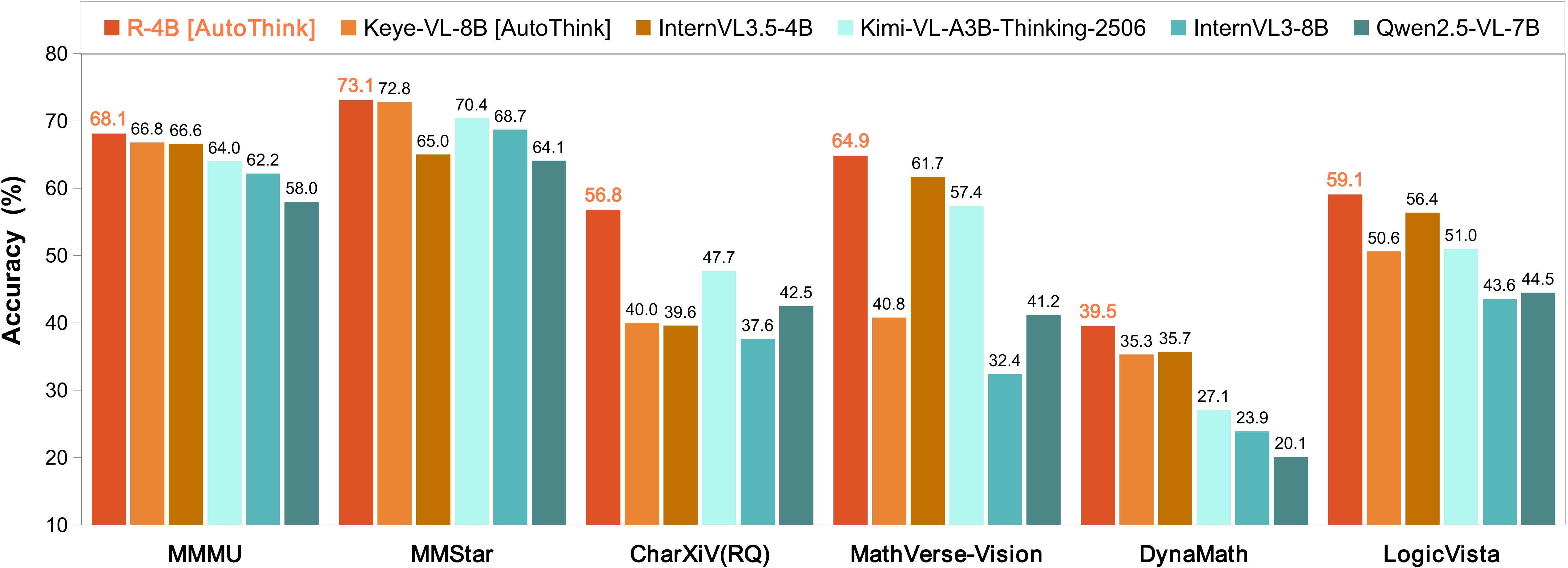}
   \caption{Comparison between \xvlrl{} and frontier open-source MLLMs, including non-thinking MLLMs~(\eg InternVL3-8B, Qwen2.5-VL 7B), thinking MLLMs~(\eg Kimi-VL-A3B-Thinking-2506) and auto-thinking MLLMs~(\eg Keye-VL-8B), on different benchmarks. }
\label{fig:performance}
\end{figure*}

\newpage
\section{Introduction}

\begin{figure*}[h!]
\centering
\includegraphics[width=1\linewidth]{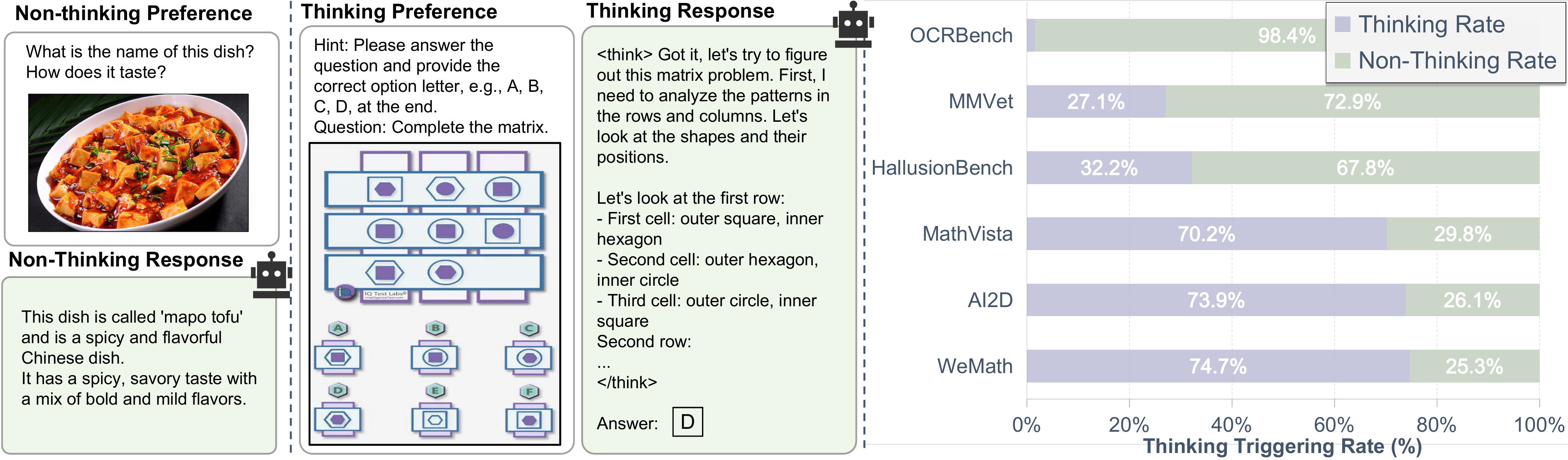}
\caption{Non-thinking and thinking mode response examples~(left); Auto-thinking triggering rates across multiple benchmarks~(right).}
\label{fig:adaptive_thinking}
\end{figure*}

Multimodal Large Language Models (MLLMs) have made significant progress in recent years, particularly through the integration of explicit, step-by-step thinking processes~\citep{guo2025seed1,coreteam2025mimovltechnicalreport,team2025kwai}.
These models employ structured token blocks to distinguish between the exploratory thinking process and the precise answer generation. 
Specifically, exploratory thinking is incorporated into the \texttt{<think>  </think>} block. It involves detailed step-by-step deduction and leverages reflection to explore alternatives or correct past reasoning.
In contrast, precise answer generation is dedicated to producing concise and clear results.
As shown in Figure~\ref{fig:adaptive_thinking}, this thinking capability has substantially improved performance on complex problems such as mathematical reasoning and scientific diagram interpretation~\citep{yue2023mmmu,mathvista,guo2025rbench}. 
However, the default always-thinking behavior leads to unnecessary computational overhead on simple problems, such as a query like ``What is the name of the dish?''.
This motivates a more intelligent auto-thinking paradigm, \ie the model can automatically decide whether to enable thinking based on the complexity of the problem.

Previous explorations into auto-thinking have shown promise but come with their own limitations. For instance, some models like Qwen3~\citep{yang2025qwen3} require users to manually enable the thinking mode within one model. To further automate the thinking process, other methods~\citep{lou2025adacot, zhan2025kat} achieve auto-thinking by relying on manually curated data or complex reward functions during RL. However, training for these methods is dependent on carefully tuned reward strategies, and they are restricted to text-only modalities.
More recent work such as Keye-VL~\citep{team2025kwai} attempts multimodal auto-thinking for the first time by constructing data with explicit complexity analysis to trigger thinking. While effective, it requires manually constructing training data with complex analysis, which makes it imprecise. Moreover, it introduces an additional token cost during inference due to the additional complexity analysis. 
These challenges demand smarter, more computationally efficient solutions for auto-thinking.

In this study, we introduce \xvl{}, a MLLM designed for general-purpose auto-thinking. It can autonomously switch between complex reasoning and direct responses directly based on the user's query. This capability is enabled by a novel training paradigm for content-aware auto-thinking.

Considering that the foundation for auto-thinking is built upon a model mastering two distinct modes for general-purpose application, \ie thinking and non-thinking, we first propose \textbf{bi-mode annealing}, which is designed to train a model that is inherently capable of both thinking and non-thinking modes in general domains. To support this annealing, we develop a bi-mode data curation strategy to carefully construct general reasoning-intensive and direct-answer datasets. This strategy separates: (a) reasoning-intensive examples requiring reasoning (e.g., diagram analysis, logical deduction), and (b) non-reasoning examples demanding direct factual responses. Both types are formatted in a unified instruction-following structure without extra complexity analysis. Then, we perform bi-mode annealing by mixing these datasets and obtain \xvlbase{}. 
This lays a solid foundation for the model's subsequent auto-thinking training in general-purpose domains.

After the annealing stage, \xvl{} possesses both thinking and non-thinking abilities. 
However, with regard to auto-thinking, it exhibits a preference for non-thinking even for complex queries, due to the uneven distribution of reasoning and non-reasoning data. This highlights a lack of judgment in mode selection.
Therefore, to further incentivize auto-thinking, we propose \textbf{Bi-mode Policy Optimization (BPO)}, a reinforcement learning algorithm tailored for auto-thinking. 
Unlike existing RL methods~\citep{zhang2025adaptthink,lou2025adacot,team2025kwai} that require complex reward functions, extensive data dependency, or are prone to hyperparameter sensitivity, BPO leverages a simple, rule-based mathematical reward. We find that the proposed approach possesses promising generalization, achieving auto-thinking in other topics as well. Specifically, the core of BPO lies in the proposed bi-mode rollouts, which incorporate both thinking and non-thinking response trajectories simultaneously. By forcing the inclusion of thinking and non-thinking modes, the model is prevented from favoring a certain mode during RL training. This mechanism enables the model to learn an adaptive policy for optimal thinking strategy selection, leading to \xvlrl{} with enhanced adaptive thinking and improved performance across modes.

Experiments demonstrate that \xvlrl{} achieves excellent performance in auto-thinking mode, achieving an optimal balance between maximizing performance and minimizing computational overhead.
In comprehensive evaluations across multiple public benchmarks, \xvlrl{} outperforms Qwen2.5-VL-7B on nearly all tasks. 
Furthermore, as shown in Figure~\ref{fig:performance}, \xvlrl{} surpasses the performance of the significantly larger Kimi-VL-Thinking-2506 (3B activated, 16B total parameters) on reasoning-intensive benchmarks. To promote further advancement in MLLMs, we open-source \xvlrl{}, which sets a new state of the art among models of comparable scale.

\section{The design of Bi-Mode Annealing}
\begin{figure*}[t]
\vspace{-0.5cm}
\centering
\includegraphics[width= 1.0\linewidth]{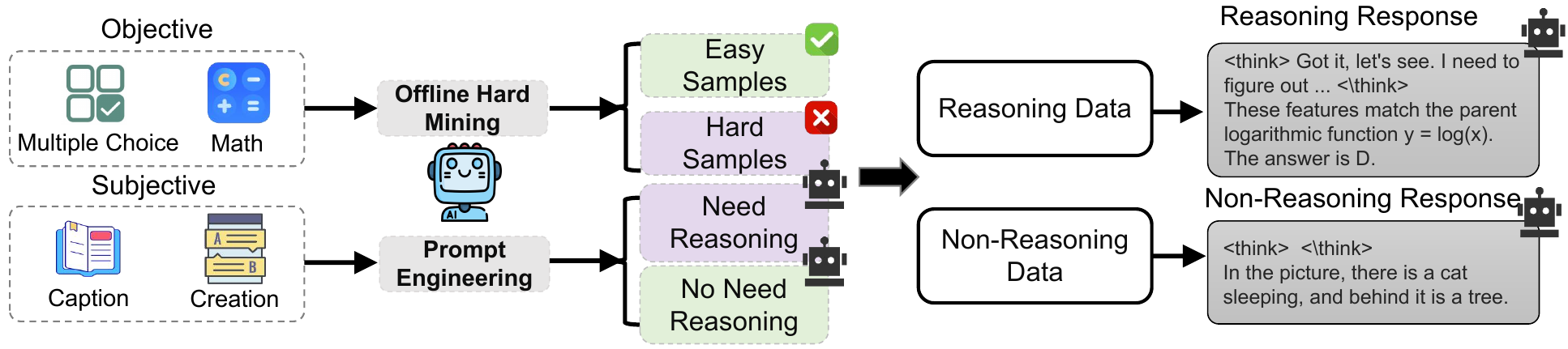}
\caption{Framework of heuristic-driven strategy for bi-mode data curation.}
\label{fig:mid-training_stage}
\vspace{-0.5cm}
\end{figure*}

\subsection{A Heuristic-Driven Strategy for Bi-Mode Data Curation}
\label{sec:reasoning_data_construct}

The goal of bi-mode annealing is to develop a model that can master two distinct response modes: thinking and non-thinking.
To achieve this, we propose a general-purpose bi-mode data curation strategy, which systematically partitions data into reasoning and non-reasoning data without laborious manual annotation.
This strategy leverages a powerful existing MLLM, Qwen2.5-32B-VL~\citep{bai2025qwen2}, to serve as a consistent annotator, partitioning data into two distinct categories: reasoning data and non-reasoning data.

As illustrated in Figure~\ref{fig:mid-training_stage}, our methodology employs two distinct heuristics tailored to the type of the query:
(i) \textbf{Difficulty-based heuristic} (for subjective queries): For queries where correctness is not easily verifiable (e.g., creative or open-ended questions), we leverage prompt engineering with an existing MLLM to assess whether they require a reasoning process based on their inherent difficulty. Queries deemed complex are labeled as reasoning-intensive.
(ii) \textbf{Performance-based heuristic} (for objective queries): For queries with verifiable answers (e.g., math or multiple choice questions), we introduce a model-based offline hard mining strategy to systematically identify difficult samples. Specifically, we generate multiple responses for each query~($N=8$); if all attempts fail~(hard samples), the query is classified as reasoning-intensive. Conversely, if the model can answer correctly~(easy samples), the query is marked as suitable for a direct answer.

\begin{wrapfigure}{r}{0.5\linewidth}
\centering
\includegraphics[width=0.9\linewidth]{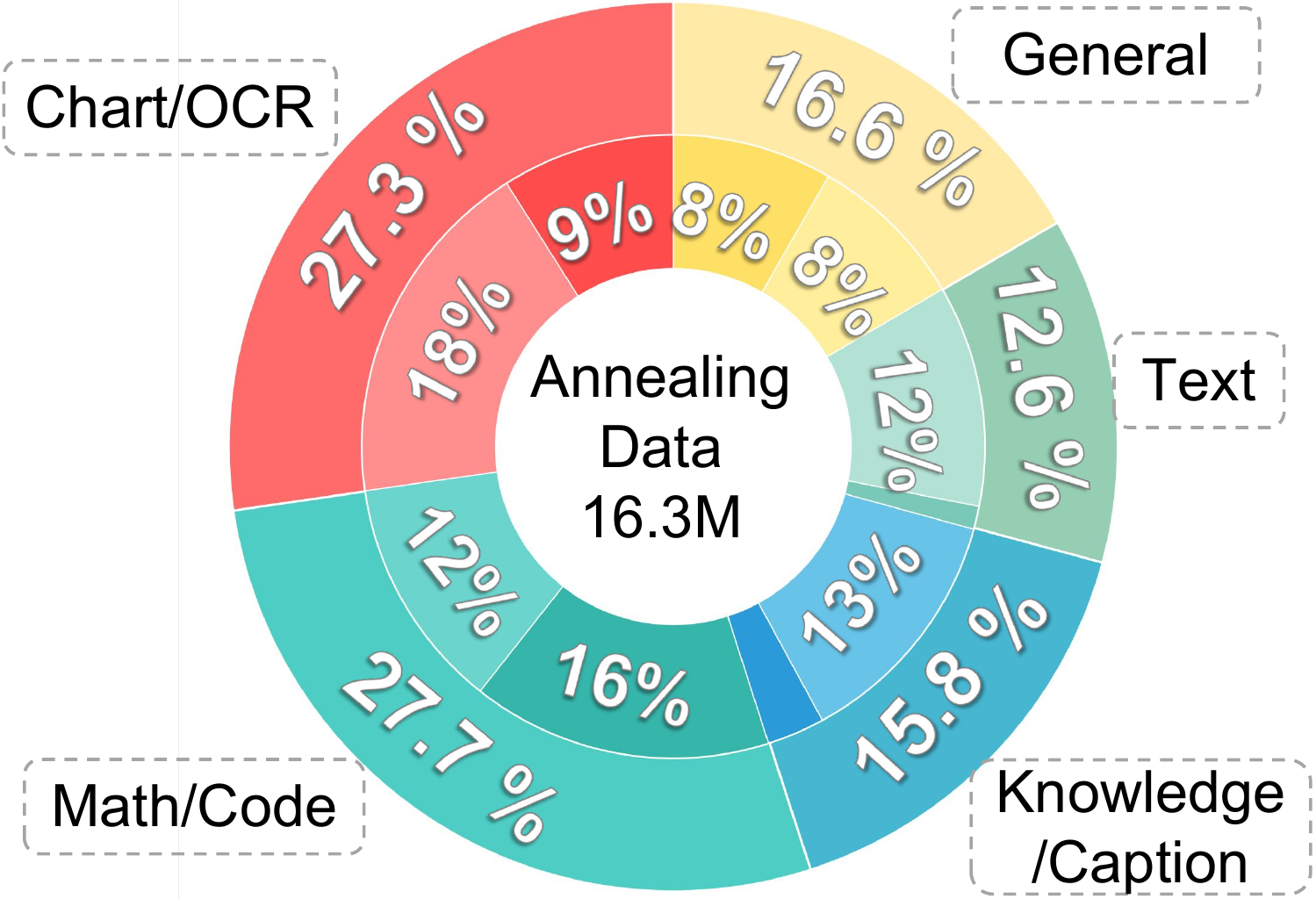}
\caption{Distributions of bi-mode data. Darker regions represent items with thinking mode, while lighter correspond to items without thinking.}
\vspace{-1cm}
\label{fig:adaptive_reasoning_rate}
\end{wrapfigure}

Subsequently, for the identified reasoning samples, we first employ the multimodal reasoning model~\citep{guo2025seed1} to extract the reasoning context. 
Then, to ensure the quality of the generated thinking processes, we filter out invalid  samples through consistency verification, keyword filtering and duplicate detection.
The resulting data proportions are shown in Figure~\ref{fig:adaptive_reasoning_rate}.
This unified data curation strategy ensures consistent annotation, eliminating inconsistencies inherent in manual labeling while adaptively addressing both question types through automated decision protocols.

\subsection{Data Formulation and Training Protocol}
With the curated bi-mode data, we reorganize the annealing dataset into the following domains: General, Math/K12, Code, Chart, OCR,  Grounding, Caption, Knowledge, and Text-Only, as detailed in Table~\ref{tab:data_distribution}. 
This structured categorization enhances the richness of the data, enabling \xvlbase{} to generalize its thinking and non-thinking capabilities across diverse scenarios.

\begin{table}[t]
\vspace{-0.5cm}
\centering
\small
\setlength{\tabcolsep}{8pt} 
\renewcommand{\arraystretch}{1.2} 
\begin{tabular}{@{}lrrrr@{}}
\toprule
\textbf{Topic Category} & \textbf{Non-reasoning Items} & \textbf{Reasoning Items} & \textbf{Total} & \textbf{Proportion} \\ 
\midrule
General & 1,351,060 & 1,365,693 & 2,716,753 & 16\% \\
Math/K12 & 1,908,486 & 1,821,412 & 4,088,776 & 23\% \\
Code & 643,323 & 161,085 & 804,408 & 5\% \\
Chart & 1,351,060 & 1,088,858 & 2,439,918 & 15\% \\
OCR & 1,366,849 & 225,146 & 1,591,995 & 10\% \\
Grounding & 280,740 & 148,710 & 429,450 & 3\% \\
Caption & 1,166,676 & 133,741 & 1,300,417 & 8\% \\
Knowledge & 928,190 & 359,778 & 1,287,967 & 8\% \\
Text-Only & 1,875,174 & 190,787 & 2,065,961 & 12\% \\
\addlinespace[3pt] 
\hline 
\addlinespace[3pt]
\textbf{Total} & \textbf{10,871,558} & \textbf{5,495,209} & \textbf{16,366,767} & \textbf{100\%} \\
\bottomrule
\end{tabular}
\caption{Data distribution across different topics in bi-mode annealing stage.}
\vspace{-0.5cm}
\label{tab:data_distribution}
\end{table}

During bi-mode annealing,  the training instances are formatted according to their designated mode:
\begin{itemize}
    \item For queries benefiting from reasoning patterns, responses include the complete reasoning process, formatted as:
    \texttt{<think>reasoning steps</think>answer}
    \item For queries suitable for direct answers, the response format maintains structural consistency but omits the thinking content, using the format:
\texttt{<think>~</think>answer}
\end{itemize}
Notably, consistent use of the \texttt{<think>} tags ensures structural uniformity. In addition, increasing the proportion of reasoning data strengthens both thinking and non-thinking capabilities in general-purpose applications. This lays the foundation for the subsequent policy optimization stage, designed to refine its auto-thinking judgment.

\section{Auto-Thinking Incentivization via Bi-mode Policy Optimization}

\begin{figure*}[t]
\centering
\vspace{-0.5cm}
\includegraphics[width= 1.0\linewidth]{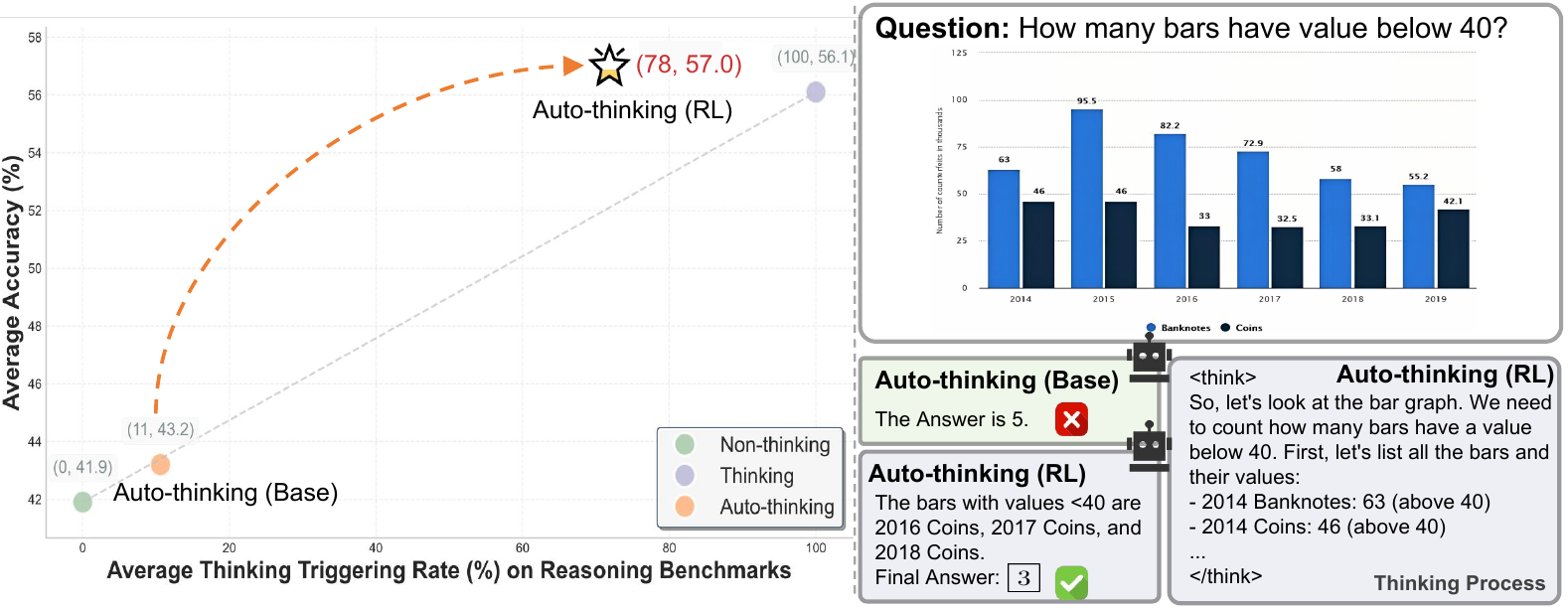}
\caption{Comparison of \xvlbase{} and \xvlrl{} in auto-thinking mode on OpenCompass Multimodal Reasoning Benchmarks. The \xvlrl{} achieves better auto-thinking inference performance.}
\label{fig:reasoning_plot}
\vspace{-0.5cm}
\end{figure*}

The bi-mode annealing stage successfully endows \xvlbase{} with the dual capabilities of reasoning (thinking) and direct response (non-thinking).
However, as illustrated in Figure~\ref{fig:reasoning_plot}, when operating in an auto-thinking inference setting in \xvlbase{}, the model exhibits a performance degradation. This phenomenon, which we term "thinking atrophy", manifests as a tendency towards non-thinking mode, even for complex queries where reasoning is essential. This instruction-following failure indicates that while the model possesses the necessary skills (\eg thinking and non-thinking capabilities), it lacks the judgment to deploy them appropriately. To bridge this gap, we transition from merely enabling these modes to actively incentivizing the optimal selection between them.

Fortunately, reinforcement learning (RL)~\citep{guo2025deepseek,yu2025dapo} presents a natural paradigm for this challenge, as it can optimize a policy based on outcome-driven rewards, thereby teaching the model to select the most effective thinking strategy. However, using vanilla RL directly will cause the model to develop thinking preferences during training, as mentioned in Section~\ref{sec:thinking_preference}. 
Some methods~\citep{lou2025adacot,zhang2025adaptthink} aim to train auto-thinking models by proposing new reward mechanisms in RL to modulate thinking preferences.
Despite their effectiveness, these approaches encounter two principal limitations:
\begin{enumerate}
\item \textbf{Complex reward engineering and data dependency:} Existing methods often rely on intricate reward functions or manually annotated data that specifies query complexity to balance exploration between thinking and non-thinking modes. This dependency introduces subjectivity, is difficult to scale, and adds significant overhead to the training pipeline.
\item \textbf{Hyperparameter sensitivity and training instability:} 
The balance between encouraging thinking and non-thinking modes in refined reward-based RL methods is typically controlled by sensitive hyperparameters.
Misconfiguration may easily lead to "mode collapse," where the policy converges to a suboptimal, single-mode strategy (either always thinking or never thinking), defeating the objective of auto-thinking.
\end{enumerate}

To overcome these challenges, we introduce Bi-mode Policy Optimization (BPO), a novel yet elegant reinforcement learning framework designed to cultivate robust auto-thinking. 
Its core principle is to learn an adaptive policy by explicitly contrasting the utility of thinking versus non-thinking pathways for the same input query.
Notably, BPO achieves a "a little goes a long way" effect. Instead of requiring complex, hand-crafted reward mechanisms or RL datasets for general domains, it leverages a simple, rule-based reward signal derived solely from the mathematical topic. We discovered that this highly specific reward mechanism possesses a remarkable universality, effectively promoting auto-thinking in diverse, non-mathematical topics as well.

\subsection{The Design of Bi-mode Policy Optimization}

\begin{figure*}[t]
\vspace{-0.5cm}
\centering
\includegraphics[width= 1\linewidth]{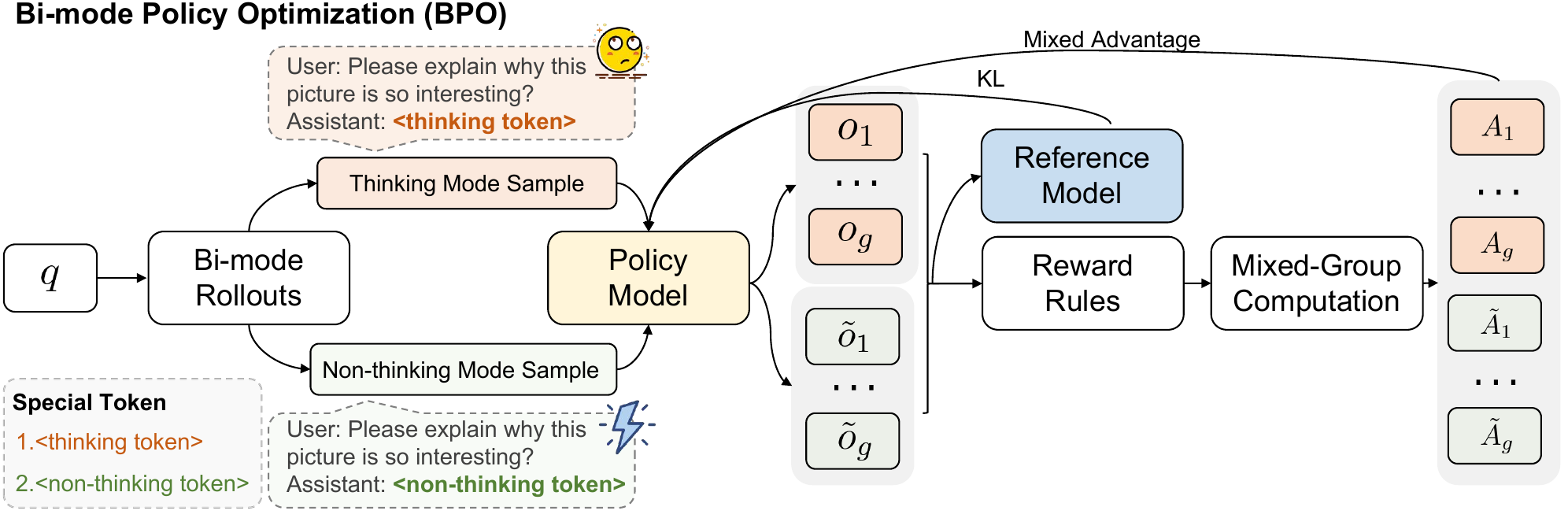}
\caption{The Bi-mode Policy Optimization (BPO) framework. For each input query, the policy model is conditioned to generate two distinct groups of responses (thinking and non-thinking).}
\label{fig:RL}
\vspace{-0.5cm}
\end{figure*}

The BPO algorithm is designed to directly optimize the model’s decision-making process, training it to select the most appropriate mode on a per-query basis.
As illustrated in Figure~\ref{fig:RL}, BPO builds upon the Group Relative Policy Optimization (GRPO) framework, but introduces a critical modification: a bi-mode sampling strategy that forces the model to generate and contrast the outcomes from both thinking modes.

Specifically, to achieve this, we deterministically generate two distinct response groups for each input prompt: a set of thinking responses, denoted as $\{o_1, \dots, o_g\}$, and a set of non-thinking responses, denoted as $\{\tilde{o}_1, \dots, \tilde{o}_g\}$. To ensure the generation of these distinct response types, we control the sampling process by conditioning on special tokens. For example, the special token sequence \textcolor[RGB]{190,90,20}{\texttt{<thinking token>}} is appended to the input prompt to trigger thinking responses, whereas the special token sequence \textcolor[RGB]{80,130,50}{\texttt{<non-thinking token>}} is used to prompt direct (non-thinking) responses. This design promotes balanced exploration across both modes by ensuring the two groups are of equal size, \ie $|\text{Group}_{\text{thinking}}| = |\text{Group}_{\text{non-thinking}}| = g$.
Formally, we propose BPO, which optimizes the policy model $\pi_\theta$ by maximizing the following objective function:
\begin{equation}
\adjustbox{max width=0.9\textwidth}{
$\displaystyle
\mathcal{J}^{\text{BPO}}(\theta) = 
\mathbb{E}_{q \sim P(Q)}
\left[
\frac{1}{2g} \sum_{k=1}^{2g} 
\min\left( R_{k}^{(o'_k)} A_{k}^{(o'_k)},\, 
\mathrm{clip}\left(R_{k}^{(o'_k)}, 1\!-\!\epsilon, 1\!+\!\epsilon\right) A_{k}^{(o'_k)} \right)
- \beta \mathbb{D}_{\text{KL}}\left(\pi_\theta(\cdot|q) \,\|\, \pi_{\text{ref}}(\cdot|q)\right)
\right]
$},
\end{equation}
where $o'_{k} = \begin{cases} {o}_{k} & \text{if } 0 < k \le g \\ \tilde{o}_{k} & \text{if } k > g \end{cases}$ represents the thinking and non-thinking rollouts respectively. $\epsilon$ and $\beta$ are hyperparameters controlling the clipping range and KL penalty strength, respectively. The policy ratio $R_k$ for each token in the generated responses and the advantage values $A_k$ are computed as in GRPO~\citep{shao2024deepseekmath}.

This deterministic grouping mechanism ensures balanced exploration between reasoning and direct answering behaviors. Through this simple yet effective reinforcement learning framework, the \xvlrl{} model achieves enhanced adaptive thinking capabilities in general-purpose domains, while simultaneously improving performance in both reasoning and non-reasoning generation.

\section{Evaluation}
This section presents a comprehensive evaluation of \xvl{}, across a diverse set of multimodal benchmarks. Specifically, we first detail the evaluation protocols in Section~\ref{sec:evaluation_setting}, ensuring reproducible comparisons. Sections~\ref{sec:general_capabilities} and \ref{sec:complex_reasoning_capabilities} explore the model's general visual understanding and advanced reasoning skills, respectively.
Furthermore, Section~\ref{sec:token_compare} compares auto-thinking token costs with other modes.
We also provide visualization cases in the supplementary material.

\subsection{Evaluation Settings}
\label{sec:evaluation_setting}
For all benchmark evaluations, we employ greedy decoding with the temperature set to 0 and a maximum generation length of 8,192 tokens for \xvl{}. To ensure fair and consistent comparisons with existing MLLMs, we adopt the widely-used VLM-EvalKit repository for evaluation. For tasks that require scoring-based evaluation, such as open-ended reasoning and complex visual understanding, we utilize Qwen3-32B as the judge model to streamline the process.
As indicated in Table~\ref{tab:results_no_gpt4o_improved}, we evaluate in three distinct modes: non-thinking (N-T) with extra token \texttt{<think>\textbackslash n\textbackslash n</think>}, thinking (T) with extra token \texttt{<think>\textbackslash n}, and auto-thinking (A-T) with extra token \texttt{<think>}. \xvlbase{} and \xvlrl{} are evaluated in the thinking and auto-thinking modes, respectively.

\begin{table*}[t]
\begin{adjustbox}{width=\textwidth, center}
\vspace{-0.5cm}
\centering
\setlength{\tabcolsep}{3pt} 

\renewcommand{\arraystretch}{1.2} 
\begin{tabular}{ll ccccc | cc}
\toprule
\multirow{3}{*}{\textbf{Capability}} & \multirow{3}{*}{\textbf{Benchmark}} & 
\textbf{Qwen2.5-VL}  & 
\textbf{InternVL3} &
\textbf{InternVL3.5} & 
\textbf{Kimi-VL-A3B} & 
\textbf{Keye-VL} & 
{\multirow{2}{*}{\textbf{\xvlbase{}}}} & 
{\multirow{2}{*}{\textbf{\xvlrl{}}}} \\
& & 
\textbf{-7B} &
\textbf{-8B} & 
\textbf{-4B} &
\textbf{-Thinking} &
\textbf{-8B} &
\\
& & 
{(N-T)} & 
{(N-T)} & 
{(T)} &
{(T)} & 
{(A-T)} & 
{(T)} & 
{(A-T)} \\
\midrule

\multirow{12}{*}{\shortstack[l]{General\\Visual QA}} 
& MMMU$_\text{val}$ & 58.6 & 62.7 & 66.6 & 64.0 & \underline{66.8} & 63.2 & \textbf{68.1} \\
& MMMU-Pro  & 34.7 & 45.6 & - & \textbf{49.2} & \underline{47.5} &46.7 & {46.5}\\  
& MMStar & 64.1 & 68.7 & 65.0 & 70.4 & \underline{72.8} & 70.8 & \textbf{73.1} \\
& MMBenchV1.1-EN$_\text{dev}$ & 82.1 & {84.7} & - & 82.6 & \textbf{89.7} & 81.9 & \underline{84.9}\\
& MMBenchV1.1-CN$_\text{dev}$ & 81.3 & 83.6 & - & 80.7 & \textbf{89.8} &83.2& \underline{84.7} \\
& MMVet & 69.7 & \underline{82.8} & 76.6 & 81.9 & 65.5 & \textbf{85.9}  & {81.9} \\
& HallusionBench & 55.7 & 49.4 & 44.8 & 57.2 & \underline{57.3} & 53.9 & \textbf{58.9} \\
& VLMs are Blind              & 37.4 & 36.8 & - & \underline{60.8} &\textbf{61.0} & 47.0 & 52.3 \\
& MMVP                        & 73.3 & 79.3 & - & \underline{80.3} & 79.0 & 79.3 &  \textbf{80.7}  \\
& VisuLogic & 20.0 & \textbf{26.1} & - & {25.0} & 21.1 & 22.5 & \underline{25.1} \\
& RealWorldQA    & 68.2 & \textbf{70.6} & 66.3 & 66.1 & 66.3 & \underline{70.5} & {69.1}\\
\midrule
\multirow{4}{*}{\shortstack[l]{Table \\\& Chart \\\& OCR}} 
& AI2D & 83.9 & 85.2 & 83.9 & 82.7 & \underline{85.8} & 84.8 & \textbf{86.2} \\
& CharXiv (DQ)  & 73.9 & 73.6 & 71.1 & 75.4 & 74.5 & \underline{82.8} & \textbf{82.9} \\
& CharXiv (RQ) & 42.5 & 37.6 & 39.6 & 47.7  &  40.0 & \underline{55.4} & \textbf{56.8} \\
& DocVQA$_\text{val}$         & \textbf{95.5} & 89.4 & \underline{92.4} & 69.0 & 86.3  & 89.6 & 91.0 \\

\midrule

\multirow{3}{*}{\shortstack[l]{Visual \\ Perception\\ \& Counting}} 
& OCRBench  & \textbf{89.7} & \underline{88.0} & 81.5 & 86.2 & 85.3 & 82.8 & 83.6 \\
& BLINK$_\text{val}$          & \underline{56.4} & 55.5 & \textbf{58.1} & 56.2 & 52.5 & 54.8 & 56.3 \\
& CountBench                  & 74.1 & 80.0 & - & \underline{91.4}  & 75.4 & \textbf{92.6} & 90.2    \\
\midrule

\multirow{7}{*}{\shortstack[l]{Math\\ \& Reasoning}} 
& MathVision                  & 26.2 & 28.8 & 26.2 & \textbf{56.8} & 42.4 & 45.7 &  \underline{47.8}  \\
& MathVista$_\text{MINI}$     & 66.8 & 70.7 & 77.1 & \textbf{80.1} & 75.2 & 76.8 & \underline{78.0} \\
& MathVerse-vision            & 41.2 & 32.4 & 61.7 & 57.4 & 40.8 & \textbf{65.0} & \underline{64.9} \\
& OlympiadBench  & 19.4 & 25.9 & - & 33.9 & 45.2 & \underline{47.0} & \textbf{49.6}  \\
& WeMath  & 37.7 & 38.5 & 50.1 & 47.0 & \textbf{58.6}& \underline{54.1} & {52.8} \\
& LogicVista & 44.5 & 43.6 & 56.4 & 51.0 & 50.6 & \underline{58.8} & \textbf{59.1} \\
& DynaMath & 20.1 & 23.9 & 35.7 & 27.1 & {35.3} & \underline{36.3} & \textbf{39.5} \\
\bottomrule
\end{tabular}
\end{adjustbox}
\caption{
Performance comparison of multimodal large language models across diverse benchmarks. 
The best and second-best results are highlighted in \textbf{bold} and \underline{underlined}, respectively. We denote thinking, non-thinking, and auto-thinking modes as T, N-T, and A-T.
Notably, Keye-VL-8B and \xvlrl{} are evaluated in A-T mode.
}
\vspace{-0.5cm}
\label{tab:results_no_gpt4o_improved}
\end{table*}


\subsection{General Capabilities}
\label{sec:general_capabilities}
\paragraph{General visual question answering.}
As shown in Table~\ref{tab:results_no_gpt4o_improved}, \xvl{} demonstrates exceptional performance on a wide array of general visual question answering benchmarks. On \texttt{MMMU$_\text{val}$}~\citep{yue2023mmmu}, a challenging multi-disciplinary benchmark, \xvlrl{} achieves a state-of-the-art score of \textbf{68.1\%}, outperforming all other models, including Keye-VL-8B (66.8\%). While Kimi-VL-A3B-Thinking leads on \texttt{MMMU-Pro}~\citep{mmmupro}, our models remain highly competitive, with \xvlbase{} (46.7\%) and \xvlrl{} (46.5\%) surpassing several larger counterparts. On \texttt{MMStar}~\citep{chen2024we}, \xvlrl{} secures the second-best position with a score of \underline{72.1\%}, closely following the top-performing Keye-VL-8B (72.8\%).
This strong performance extends to standard \texttt{MMBench}~\citep{liu2023mmbench} evaluations. On both \texttt{MMBenchV1.1-EN$_\text{dev}$} and \texttt{MMBenchV1.1-CN$_\text{dev}$}, \xvlrl{} consistently ranks second with scores of \underline{84.9\%} and \underline{84.7\%}, respectively, reaffirming its robust cross-lingual understanding. Notably, on \texttt{MMVet}~\citep{yu2023mm}, our base model \xvlbase{} achieves a remarkable score of \textbf{85.9\%}, significantly surpassing all competitors. 
Furthermore, \xvl{} is good at tasks with visual illusions and logical fallacies. It achieves the top score of \textbf{58.9} on \texttt{HallusionBench}~\citep{guan2023hallusionbench} and \textbf{80.7\%} on \texttt{MMVP}~\citep{tong2024eyes}, setting a new standard on both benchmarks. On the logic-intensive \texttt{VisuLogic}~\citep{xu2025visulogic} benchmark, \xvlrl{} attains a competitive second-place score of \underline{25.1\%}. Similarly, on \texttt{RealWorldQA}, \xvlbase{} is the second-best performer with \underline{70.5\%}. These results collectively underscore \xvl{}'s superior general visual understanding and reasoning capabilities.

\paragraph{Document and chart understanding.}
\xvl{} demonstrates outstanding proficiency in interpreting structured visual content like documents, diagrams, and charts. On \texttt{AI2D}~\citep{kembhavi2016diagram}, a diagram understanding benchmark, \xvlrl{} achieves the highest score of \textbf{86.2\%}, showcasing its advanced spatial and semantic comprehension. The model's dominance is particularly evident in chart analysis. On \texttt{CharXiv}~\citep{wang2024charxiv}, \xvlrl{} not only leads in Descriptive Questions~(DQ) with a score of \textbf{82.9\%} but also establishes a significant lead in Reasoning Questions~(RQ) with a score of \textbf{56.8\%}. This is a substantial improvement of over 9 percentage points compared to the Kimi-VL-A3B-Thinking (47.7\%), highlighting its superior ability to reason over complex chart semantics.
On \texttt{DocVQA$_\text{val}$}~\citep{mathew2021docvqa}, \xvlrl{} obtains a strong score of 91.0\%.
This performance confirms \xvl{}'s exceptionally robust capabilities in both precise text extraction and document structure interpretation.

\paragraph{Visual perception and counting.}
In tasks requiring fine-grained visual perception, \xvl{} delivers highly competitive results. On \texttt{BLINK$_\text{val}$}~\citep{fu2024blink}, a benchmark for referring expression comprehension, \xvlrl{} scores \underline{56.3\%}, effectively tying for the top position with Qwen2.5-VL (56.4\%) and outperforming other models. The most impressive result in this category comes from \texttt{CountBench}~\citep{paiss2023teaching}, where our base model, \xvlbase{}, sets the highest score with \textbf{92.6\%}, demonstrating exceptional object counting abilities. \xvlrl{} also performs strongly with a score of 90.2\%. On \texttt{OCRBench}~\citep{liu2024ocrbenchhiddenmysteryocr}, \xvlrl{} achieves a robust score of 83.6\%, proving its solid text recognition capabilities in diverse scenes.

\subsection{Complex Reasoning Capabilities}
\label{sec:complex_reasoning_capabilities}
\xvl{} exhibits exceptional strength in complex mathematical and logical reasoning tasks. Notably, our models secure the top two positions on several challenging math-focused evaluations. On \texttt{MathVerse-vision}~\citep{zhang2024mathverse}, \xvlbase{} (\textbf{65.0\%}) and \xvlrl{} (\underline{64.9\%}) dramatically outperform all prior models, with the next-best competitor lagging at 57.4\%. A similar dominance is observed on \texttt{OlympiadBench}~\citep{he2024olympiadbench}, where \xvlrl{} (\textbf{49.6\%}) and \xvlbase{} (\underline{47.0\%}) take the lead. The trend continues on \texttt{LogicVista}~\citep{xiao2024logicvista} (\xvlrl{}: \textbf{59.1\%}, \xvlbase{}: \underline{58.8\%}) and \texttt{DynaMath}~\citep{zou2024dynamath} (\xvlrl{}: \textbf{39.5\%}, \xvlbase{}: \underline{36.3\%}), where our models again claim the top two spots, underscoring their advanced logical deduction skills.
On other prominent math benchmarks, \xvl{} remains a top contender. It achieves the second-best scores on \texttt{MathVision}~\citep{mathvision} (\underline{47.8\%}) and \texttt{MathVista$_\text{MINI}$}~\citep{mathvista} (\underline{78.0\%}), trailing only Kimi-VL-A3B-Thinking but significantly surpassing other baselines. 
These results demonstrate that our advanced training strategies empower \xvl{} to compete with larger MLLMs and \xvlrl{} sets a new standard for 4B MLLMs.

\begin{figure*}[t]
\centering
\vspace{-0.5cm}
\includegraphics[width= 1\linewidth]{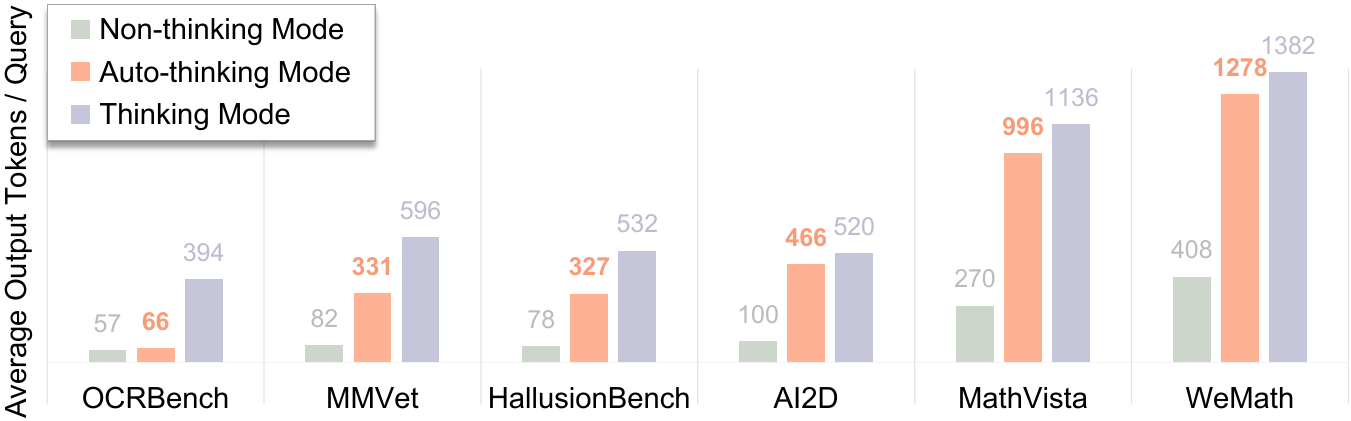}
\caption{Comparison of average output tokens per query across non-thinking, auto-thinking, and thinking modes on different benchmarks. The auto-thinking mode achieves a trade-off between efficiency and performance.
}
\label{fig:tokens_static}
\vspace{-0.5cm}
\end{figure*}

\subsection{Token Consumption Across Different Modes}
\label{sec:token_compare}
As illustrated in Figure~\ref{fig:tokens_static}, we analyze the average output tokens per query of \xvlrl{}, to verify the effectiveness of our BPO method. This analysis provides empirical evidence for the generalization capability of the auto-thinking learned via BPO. Although trained with a reward signal exclusively from the mathematical domain, \xvlrl{} demonstrates an understanding of task complexity across a diverse set of general-purpose benchmarks.
For simpler tasks such as those in \texttt{OCRBench}, the auto-thinking mode generates only 66 tokens, a volume comparable to the non-thinking mode (57 tokens) and significantly lower than the full thinking mode (394 tokens). Besides, its performance of 83.6\% matches the non-thinking mode (83.6\%) and even surpasses the thinking mode (82.6\%). This indicates that for straightforward queries, our model conserves computational resources while maintaining performance.
Meanwhile, when confronted with reasoning-intensive benchmarks like \texttt{MathVista} and \texttt{WeMath}, the auto-thinking mode dynamically increases its token output to 996 and 1278, respectively. These figures closely approach the token counts of the thinking mode (1136 and 1382 tokens). At the same time, this enables our model to achieve performance of 78.0\% on \texttt{MathVista} and 52.8\% on \texttt{WeMath}, which are substantially higher than the non-thinking mode's scores (71.5\% and 46.6\%) and competitive with the thinking mode (79.7\% and 55.8\%). In conclusion, these findings validate that our BPO is not merely a thinking switch but a truly intelligent and generalizable policy. It effectively discerns task complexity across varied domains, striking an optimal trade-off between performance and efficiency.
\section{Analysis}

\paragraph{Training data ablation in annealing stage.}
We performed an ablation study to determine the optimal bi-mode annealing strategy, by focusing on different data compositions. As detailed in Table~\ref{tab:ablation_training_data}, results demonstrate that a mixed-data approach combined with a thinking mode is markedly superior. This strategy achieves an average performance of 69.5\%, surpassing models trained on only reasoning data (+4.1\%) and a two-stage curriculum (+2.6\%).
In addition, while the Only-R strategy enhances performance on specific complex tasks (e.g., +8.7\% on MathVision), it fails to generalize tasks, leading to a lower overall score. The Mixed-R strategy, however, successfully balances specialized reasoning with general capabilities. This suggests that co-training on both data types prevents catastrophic forgetting of general skills while effectively instilling complex reasoning abilities.

\begin{figure}[tbp]
\centering
\begin{minipage}{0.48\linewidth}
\centering
\includegraphics[width=\linewidth]{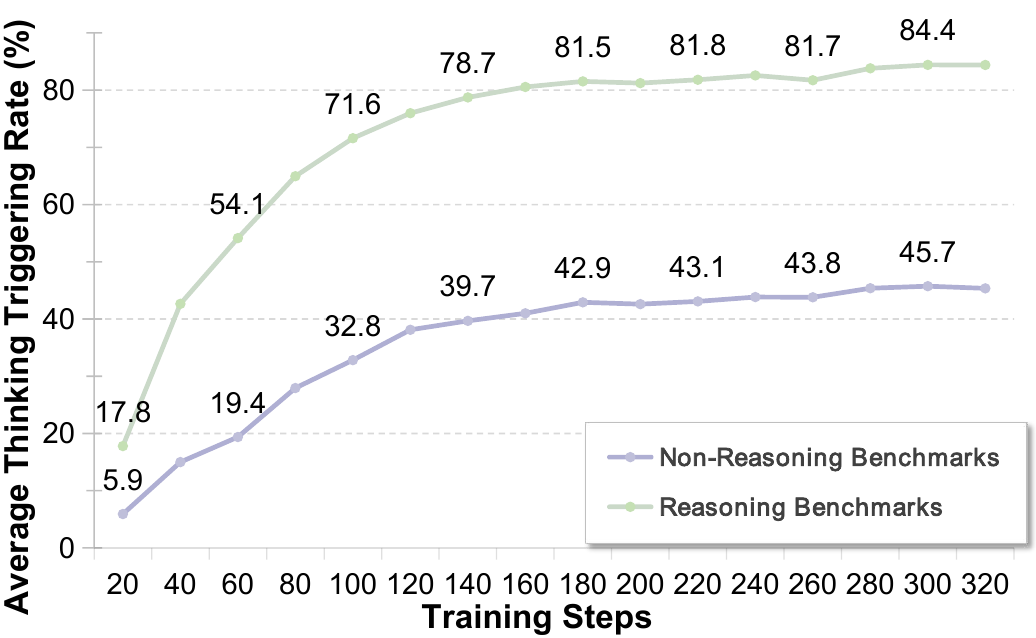}
\caption{Average thinking triggering rate (\%) across training steps for non-reasoning benchmarks, and reasoning benchmarks.}
\label{fig:annealing_stage_rate}
\end{minipage}
\hfill
\begin{minipage}{0.48\linewidth}
\centering
\includegraphics[width=\linewidth]{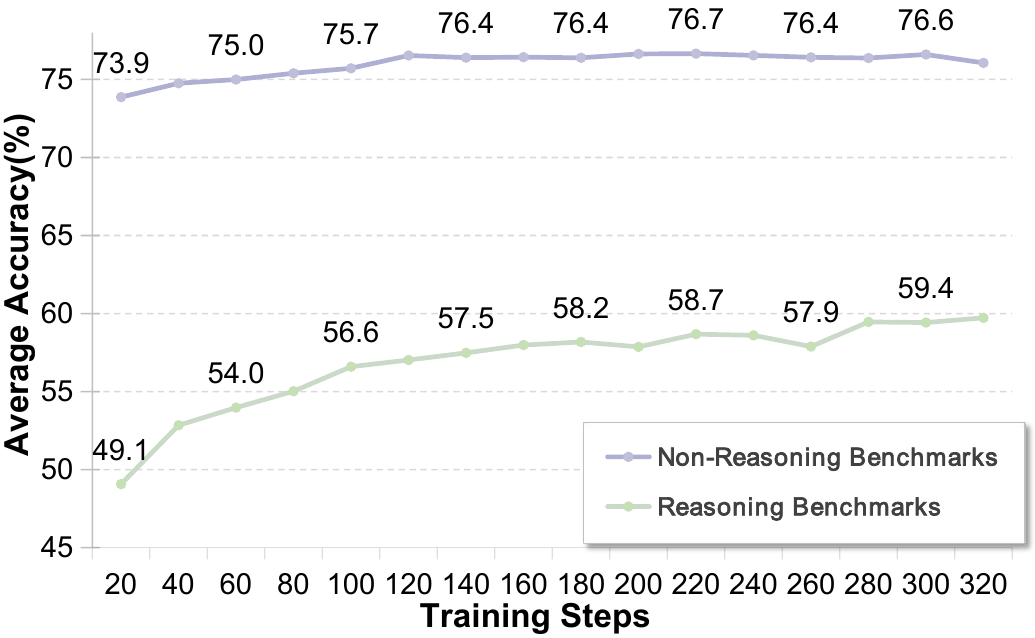}
\caption{Average accuracy (\%) across RL training steps for non-reasoning benchmarks, and reasoning benchmarks.}
\label{fig:annealing_stage_acc}
\end{minipage}
\end{figure}

\begin{table}[tbp]
\centering
\vspace{-0.5cm}
\setlength{\tabcolsep}{4pt} 
\renewcommand{\arraystretch}{1.3}
\resizebox{\linewidth}{!}{
\begin{tabular}{lcc*{7}{S[table-format=2.1]}|S[table-format=2.1]}
\toprule
{\textbf{Strategy}} & {\textbf{Data}} & {\textbf{Mode}} & {\textbf{MMMU}} & {\textbf{MMStar}} & {\textbf{AI2D}} & {\textbf{OCRBench}} & {\textbf{MathVista}} & {\textbf{MathVision}} & {\textbf{LogicVista}} & {\textbf{Average}} \\

\midrule
Non-R             & ~16.3M           & N-T      & 60.4 & 70.4 & 83.2 & {83.3} & 71.1 & 33.2 & 49.0 & 64.4 \\
Only-R            & ~5.5M            & T          & 62.8 & 65.7 & 78.9 & 82.1 & 73.6 & 41.9 & 52.6 & 65.4 \\
Non-R $\to$ R & ~10.8M $\to$ ~5.5M   & T         & 63.8 & 68.7 & 80.4 & 82.7 & 74.9 & 43.7 & 54.4 & 66.9 \\
Mixed-R       & ~16.3M           & N-T      & 56.3 & 67.5 & {84.5} & 82.7 & 65.5 & 29.9 & 47.7 & 62.0 \\
Mixed-R       & ~16.3M           & T          & {64.6} & {73.1} & {84.5} & 82.8 & 76.8 & 45.7 & {58.8} & \textbf{69.5} \\
\bottomrule
\multicolumn{10}{@{}l}{\small\textit{*Note:} Non-R $\to$ Reasoning strategy resulting model possesses reasoning capabilities only.} \\
\end{tabular}
}
\caption{ Ablation study on training strategies during the annealing stage. We compare four data strategies: using only non-reasoning data (Non-R), only reasoning data (Only-R), a two-stage curriculum (Non-R $\to$ R), and a mix of both reasoning data and non-reasoning data~(Mixed-R). Performance is evaluated in either non-thinking (N-T) or thinking (T) inference mode.}
\vspace{-0.2cm}
\label{tab:ablation_training_data}
\end{table}
\paragraph{Thinking triggering rate during reinforcement learning.}
To dissect the learning process of BPO, we tracked the evolution of the thinking triggering rate across benchmarks with distinct complexity profiles. As illustrated in Figure~\ref{fig:annealing_stage_rate}, on reasoning-intensive benchmarks (e.g., \texttt{MathVision}, \texttt{MathVista}), the model rapidly learns to activate its thinking mode, with the trigger rate showing a steep initial climb before stabilizing at a high level. Conversely, on non-reasoning benchmarks (e.g., \texttt{OCRBench}, \texttt{HallusionBench}), the trigger rate shows only a marginal and slow increase.
This differential learning processing is the direct consequence of our BPO mechanism. The model quickly discovers that applying the thinking mode to reasoning tasks yields a high reward signal, thus reinforcing this behavior. Simultaneously, it learns that `thinking' on simpler, non-reasoning tasks offers few rewards. 
The benefit of our RL policy is reflected in the performance gains. As shown in Figure~\ref{fig:annealing_stage_acc}, BPO achieves a substantial +10.3\% improvement on reasoning tasks, a gain that far surpasses its impact on non-reasoning tasks. This directly validates BPO's effectiveness in mitigating "thinking atrophy" by incentivizing the appropriate auto-thinking.

\paragraph{Comparison between \xvlbase{} and \xvlrl{}.}
\label{sec:compare_rl_and_base}

Table~\ref{tab:performance_comparison} presents a comprehensive comparison between \xvlbase{} and \xvlrl{} across multiple reasoning benchmarks under non-thinking, thinking, and auto-thinking modes. The results demonstrate that RL consistently enhances performance in both non-thinking and thinking capabilities.
In \textit{non-thinking mode}, where models generate answers directly without explicit reasoning, \xvlrl{} achieves a substantial improvement over \xvlbase{}, raising the average score from 42.0\% to 49.9\%, outperforming the base model on all benchmarks. 
This significant gain underscores the effectiveness of RL for direct response generation, thereby strengthening non-reasoning capabilities.
Besides, in thinking mode, \xvlrl{} retains its edge, pushing the average from 56.1\% to 58.1\%, showing greater benefits from reasoning. 
Overall, \xvlrl{} consistently outperforms \xvlbase{} in both modes, confirming our RL method boosts both thinking and non-thinking performance.

\begin{table}[tbp]
\centering
\setlength{\tabcolsep}{4pt} 
\resizebox{\linewidth}{!}{
\begin{tabular}{l c c c c c c c| c}
\toprule
\textbf{Model} & \textbf{Mode} & {\textbf{MathVista}} & {\textbf{MathVision}} & {\textbf{MathVerse}} & {\textbf{DynaMath}} & {\textbf{WeMath}} & {\textbf{LogicVista}} & {\textbf{Average}} \\
\midrule
\xvlbase{} & N-T & 65.5 & 29.9 & 48.4 & 28.5 & 32.0 & 47.7 & 42.0 \\
\xvlrl{}       & N-T & 71.5 & 35.0 & 61.5 & 31.7 & 46.6 & 52.8 & 49.9 \\
\xvlbase{}       & A-T & 65.7 & 30.8 & 48.9 & 30.0 & 34.3 & 49.4 & 43.2 \\
\xvlrl{}       & A-T & 78.0 & 47.8 & 64.9 & 39.5 & 52.8 & 59.1 & 57.0 \\
\xvlbase{} & T     & 76.8 & 45.7 & 65.0 & 36.3 & 54.2 & 58.8 & 56.1 \\
\xvlrl{}          & T     & 79.7 & 47.9 & 67.1 & 39.5 & 55.8 & 58.6 & \textbf{58.1} \\
\bottomrule
\end{tabular}
}
\caption{Performance comparison of \xvlbase{} and \xvlrl{} models across reasoning benchmarks.}
\label{tab:performance_comparison}
\end{table}

\begin{wrapfigure}{r}{0.5\linewidth}
\centering
\includegraphics[width=1.0\linewidth]{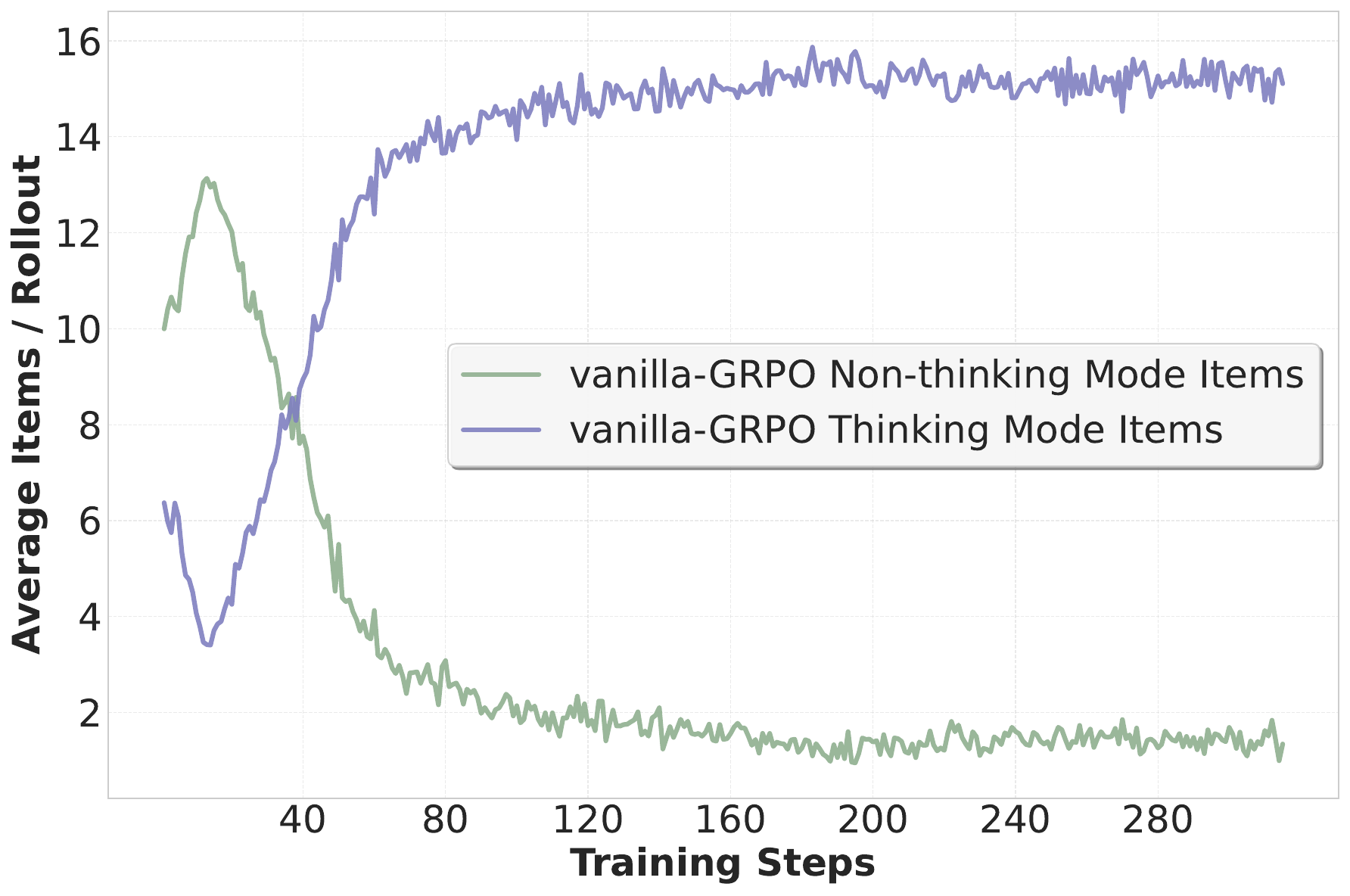}
\caption{The average thinking and non-thinking items across training steps on vanilla GRPO.}
\label{fig:annealing_stage_item}
\vspace{-0.5cm}
\end{wrapfigure}
\paragraph{Thinking preference dilemma of vanilla GRPO.}
\label{sec:thinking_preference}
As shown in Figure~\ref{fig:annealing_stage_item}, we compare the average items per rollout for vanilla GRPO, considering both thinking and non-thinking scenarios, across the reinforcement learning training steps.
For GRPO, the non-thinking items increase initially but then drop and remain at a low average correct items per rollout, which falls into the thinking preference dilemma.
Overall, BPO demonstrates better performance in both thinking and non-thinking modes across the training steps, indicating that BPO is more effective in handling adaptive reasoning in the RL training stage compared to GRPO and avoid the preference dilemma.

\section{Related Work}

Recently, existing methods to enhance the efficiency of LLMs have centered on manually enabling the thinking mode in model responses~\citep{yang2025qwen3}. 
To automate this process, some methods incorporate RL with extra thinking preference rewards to encourage models to think on their own~\citep{tu2025learning,zhang2025adaptthink}.
Other approaches further fine-tune models on datasets containing both thinking and non-thinking responses, which are obtained from prompt engineering and existing reasoning models~\citep{lou2025adacot,zhan2025kat}.
In the multimodal domain, Keye-VL~\citep{team2025kwai} addresses auto-thinking using Mix-Mode RL.
Inspired by the above methods, by introducing bi-mode annealing and BPO strategy, \xvl{} enables efficient, content-dependent auto-thinking by a single model, filling a crucial gap towards efficient and auto-thinking MLLMs.

\section{Conclusion}

In this work, we introduce \xvl{}, a multimodal large language model designed to address the critical trade-off between complex reasoning and inference efficiency. \xvl{} features an "auto-thinking" mechanism that dynamically switches between thinking and direct answering modes. Our method first equips \xvlbase{} with thinking and direct answering capabilities through a bi-mode annealing stage. Subsequently, we employ bi-mode policy optimization, a reinforcement learning approach that utilizes hybrid mixed-policy rollouts to mitigate "thinking collapse" and learn an optimal policy for mode selection in general domains. \xvlrl{} achieves state-of-the-art performance on reasoning-intensive benchmarks, outperforming comparable models and matching the larger Kimi-VL-A3B-Thinking-2506. This research demonstrates a practical and effective pathway toward developing more intelligent and resource-efficient MLLMs.

\newpage

\bibliography{iclr2025_conference}

\begin{thebibliography}{39}
\providecommand{\natexlab}[1]{#1}
\providecommand{\url}[1]{\texttt{#1}}
\expandafter\ifx\csname urlstyle\endcsname\relax
  \providecommand{\doi}[1]{doi: #1}\else
  \providecommand{\doi}{doi: \begingroup \urlstyle{rm}\Url}\fi

\bibitem[Bai et~al.(2025)Bai, Chen, Liu, Wang, Ge, Song, Dang, Wang, Wang, Tang, et~al.]{bai2025qwen2}
Shuai Bai, Keqin Chen, Xuejing Liu, Jialin Wang, Wenbin Ge, Sibo Song, Kai Dang, Peng Wang, Shijie Wang, Jun Tang, et~al.
\newblock Qwen2. 5-vl technical report.
\newblock \emph{arXiv preprint arXiv:2502.13923}, 2025.

\bibitem[Chen et~al.(2024{\natexlab{a}})Chen, Li, Dong, Zhang, Zang, Chen, Duan, Wang, Qiao, Lin, et~al.]{chen2024we}
Lin Chen, Jinsong Li, Xiaoyi Dong, Pan Zhang, Yuhang Zang, Zehui Chen, Haodong Duan, Jiaqi Wang, Yu~Qiao, Dahua Lin, et~al.
\newblock Are we on the right way for evaluating large vision-language models?
\newblock \emph{arXiv:2403.20330}, 2024{\natexlab{a}}.

\bibitem[Chen et~al.(2024{\natexlab{b}})Chen, Wang, Cao, Liu, Gao, Cui, Zhu, Ye, Tian, Liu, et~al.]{chen2024expanding}
Zhe Chen, Weiyun Wang, Yue Cao, Yangzhou Liu, Zhangwei Gao, Erfei Cui, Jinguo Zhu, Shenglong Ye, Hao Tian, Zhaoyang Liu, et~al.
\newblock Expanding performance boundaries of open-source multimodal models with model, data, and test-time scaling.
\newblock \emph{arXiv preprint arXiv:2412.05271}, 2024{\natexlab{b}}.

\bibitem[Fu et~al.(2024)Fu, Hu, Li, Feng, Wang, Lin, Roth, Smith, Ma, and Krishna]{fu2024blink}
Xingyu Fu, Yushi Hu, Bangzheng Li, Yu~Feng, Haoyu Wang, Xudong Lin, Dan Roth, Noah~A Smith, Wei-Chiu Ma, and Ranjay Krishna.
\newblock Blink: Multimodal large language models can see but not perceive.
\newblock In \emph{European Conference on Computer Vision}, pp.\  148--166. Springer, 2024.

\bibitem[Guan et~al.(2023)Guan, Liu, Wu, Xian, Li, Liu, Wang, Chen, Huang, Yacoob, Manocha, and Zhou]{guan2023hallusionbench}
Tianrui Guan, Fuxiao Liu, Xiyang Wu, Ruiqi Xian, Zongxia Li, Xiaoyu Liu, Xijun Wang, Lichang Chen, Furong Huang, Yaser Yacoob, Dinesh Manocha, and Tianyi Zhou.
\newblock Hallusionbench: An advanced diagnostic suite for entangled language hallucination \& visual illusion in large vision-language models.
\newblock \emph{arXiv:2310.14566}, 2023.

\bibitem[Guo et~al.(2025{\natexlab{a}})Guo, Yang, Zhang, Song, Zhang, Xu, Zhu, Ma, Wang, Bi, et~al.]{guo2025deepseek}
Daya Guo, Dejian Yang, Haowei Zhang, Junxiao Song, Ruoyu Zhang, Runxin Xu, Qihao Zhu, Shirong Ma, Peiyi Wang, Xiao Bi, et~al.
\newblock Deepseek-r1: Incentivizing reasoning capability in llms via reinforcement learning.
\newblock \emph{arXiv preprint arXiv:2501.12948}, 2025{\natexlab{a}}.

\bibitem[Guo et~al.(2025{\natexlab{b}})Guo, Wu, Zhu, Leng, Shi, Chen, Fan, Wang, Jiang, Wang, et~al.]{guo2025seed1}
Dong Guo, Faming Wu, Feida Zhu, Fuxing Leng, Guang Shi, Haobin Chen, Haoqi Fan, Jian Wang, Jianyu Jiang, Jiawei Wang, et~al.
\newblock Seed1. 5-vl technical report.
\newblock \emph{arXiv preprint arXiv:2505.07062}, 2025{\natexlab{b}}.

\bibitem[Guo et~al.(2025{\natexlab{c}})Guo, Chu, Yang, Mo, Shen, Li, Lin, Zhang, Chen, Zhang, et~al.]{guo2025rbench}
Meng-Hao Guo, Xuanyu Chu, Qianrui Yang, Zhe-Han Mo, Yiqing Shen, Pei-lin Li, Xinjie Lin, Jinnian Zhang, Xin-Sheng Chen, Yi~Zhang, et~al.
\newblock Rbench-v: A primary assessment for visual reasoning models with multi-modal outputs.
\newblock \emph{arXiv preprint arXiv:2505.16770}, 2025{\natexlab{c}}.

\bibitem[He et~al.(2024)He, Luo, Bai, Hu, Thai, Shen, Hu, Han, Huang, Zhang, et~al.]{he2024olympiadbench}
Chaoqun He, Renjie Luo, Yuzhuo Bai, Shengding Hu, Zhen~Leng Thai, Junhao Shen, Jinyi Hu, Xu~Han, Yujie Huang, Yuxiang Zhang, et~al.
\newblock Olympiadbench: A challenging benchmark for promoting agi with olympiad-level bilingual multimodal scientific problems.
\newblock \emph{arXiv preprint arXiv:2402.14008}, 2024.

\bibitem[Kembhavi et~al.(2016)Kembhavi, Salvato, Kolve, Seo, Hajishirzi, and Farhadi]{kembhavi2016diagram}
Aniruddha Kembhavi, Mike Salvato, Eric Kolve, Minjoon Seo, Hannaneh Hajishirzi, and Ali Farhadi.
\newblock A diagram is worth a dozen images.
\newblock In \emph{European conference on computer vision}, pp.\  235--251. Springer, 2016.

\bibitem[Li et~al.(2024)Li, Zhang, Guo, Zhang, Li, Zhang, Zhang, Zhang, Li, Liu, et~al.]{li2024llava}
Bo~Li, Yuanhan Zhang, Dong Guo, Renrui Zhang, Feng Li, Hao Zhang, Kaichen Zhang, Peiyuan Zhang, Yanwei Li, Ziwei Liu, et~al.
\newblock Llava-onevision: Easy visual task transfer.
\newblock \emph{arXiv preprint arXiv:2408.03326}, 2024.

\bibitem[Liu et~al.(2024)Liu, Li, Li, and Lee]{liu2024improved}
Haotian Liu, Chunyuan Li, Yuheng Li, and Yong~Jae Lee.
\newblock Improved baselines with visual instruction tuning.
\newblock In \emph{Proceedings of the IEEE/CVF conference on computer vision and pattern recognition}, pp.\  26296--26306, 2024.

\bibitem[Liu et~al.(2023{\natexlab{a}})Liu, Duan, Zhang, Li, Zhang, Zhao, Yuan, Wang, He, Liu, et~al.]{liu2023mmbench}
Yuan Liu, Haodong Duan, Yuanhan Zhang, Bo~Li, Songyang Zhang, Wangbo Zhao, Yike Yuan, Jiaqi Wang, Conghui He, Ziwei Liu, et~al.
\newblock Mmbench: Is your multi-modal model an all-around player?
\newblock \emph{arXiv preprint arXiv:2307.06281}, 2023{\natexlab{a}}.

\bibitem[Liu et~al.(2023{\natexlab{b}})Liu, Li, Huang, Yang, Yu, Li, Yin, lin Liu, Jin, and Bai]{liu2024ocrbenchhiddenmysteryocr}
Yuliang Liu, Zhang Li, Mingxin Huang, Biao Yang, Wenwen Yu, Chunyuan Li, Xucheng Yin, Cheng lin Liu, Lianwen Jin, and Xiang Bai.
\newblock Ocrbench: On the hidden mystery of ocr in large multimodal models.
\newblock \emph{arXiv:2305.07895}, 2023{\natexlab{b}}.

\bibitem[Liu et~al.(2025)Liu, Dong, Wang, Liu, Hu, Lu, and Rao]{liu2025ola}
Zuyan Liu, Yuhao Dong, Jiahui Wang, Ziwei Liu, Winston Hu, Jiwen Lu, and Yongming Rao.
\newblock Ola: Pushing the frontiers of omni-modal language model.
\newblock \emph{arXiv preprint arXiv:2502.04328}, 2025.

\bibitem[Lou et~al.(2025)Lou, Sun, Liang, Qu, Shen, Wang, Li, Yang, and Wu]{lou2025adacot}
Chenwei Lou, Zewei Sun, Xinnian Liang, Meng Qu, Wei Shen, Wenqi Wang, Yuntao Li, Qingping Yang, and Shuangzhi Wu.
\newblock Adacot: Pareto-optimal adaptive chain-of-thought triggering via reinforcement learning.
\newblock \emph{arXiv preprint arXiv:2505.11896}, 2025.

\bibitem[Lu et~al.(2024)Lu, Bansal, Xia, Liu, Li, Hajishirzi, Cheng, Chang, Galley, and Gao]{mathvista}
Pan Lu, Hritik Bansal, Tony Xia, Jiacheng Liu, Chunyuan Li, Hannaneh Hajishirzi, Hao Cheng, Kai{-}Wei Chang, Michel Galley, and Jianfeng Gao.
\newblock Mathvista: Evaluating mathematical reasoning of foundation models in visual contexts.
\newblock In \emph{ICLR}, 2024.

\bibitem[Mathew et~al.(2021)Mathew, Karatzas, and Jawahar]{mathew2021docvqa}
Minesh Mathew, Dimosthenis Karatzas, and CV~Jawahar.
\newblock Docvqa: A dataset for vqa on document images.
\newblock In \emph{Proceedings of the IEEE/CVF winter conference on applications of computer vision}, pp.\  2200--2209, 2021.

\bibitem[Paiss et~al.(2023)Paiss, Ephrat, Tov, Zada, Mosseri, Irani, and Dekel]{paiss2023teaching}
Roni Paiss, Ariel Ephrat, Omer Tov, Shiran Zada, Inbar Mosseri, Michal Irani, and Tali Dekel.
\newblock Teaching clip to count to ten.
\newblock In \emph{Proceedings of the IEEE/CVF International Conference on Computer Vision}, pp.\  3170--3180, 2023.

\bibitem[Shao et~al.(2024)Shao, Wang, Zhu, Xu, Song, Bi, Zhang, Zhang, Li, Wu, et~al.]{shao2024deepseekmath}
Zhihong Shao, Peiyi Wang, Qihao Zhu, Runxin Xu, Junxiao Song, Xiao Bi, Haowei Zhang, Mingchuan Zhang, YK~Li, Yang Wu, et~al.
\newblock Deepseekmath: Pushing the limits of mathematical reasoning in open language models.
\newblock \emph{arXiv preprint arXiv:2402.03300}, 2024.

\bibitem[Tong et~al.(2024)Tong, Liu, Zhai, Ma, LeCun, and Xie]{tong2024eyes}
Shengbang Tong, Zhuang Liu, Yuexiang Zhai, Yi~Ma, Yann LeCun, and Saining Xie.
\newblock Eyes wide shut? exploring the visual shortcomings of multimodal llms.
\newblock In \emph{Proceedings of the IEEE/CVF Conference on Computer Vision and Pattern Recognition}, pp.\  9568--9578, 2024.

\bibitem[Tschannen et~al.(2025)Tschannen, Gritsenko, Wang, Naeem, Alabdulmohsin, Parthasarathy, Evans, Beyer, Xia, Mustafa, et~al.]{tschannen2025siglip}
Michael Tschannen, Alexey Gritsenko, Xiao Wang, Muhammad~Ferjad Naeem, Ibrahim Alabdulmohsin, Nikhil Parthasarathy, Talfan Evans, Lucas Beyer, Ye~Xia, Basil Mustafa, et~al.
\newblock Siglip 2: Multilingual vision-language encoders with improved semantic understanding, localization, and dense features.
\newblock \emph{arXiv preprint arXiv:2502.14786}, 2025.

\bibitem[Tu et~al.(2025)Tu, Lin, Zhang, Tian, Li, Lan, and Zhao]{tu2025learning}
Songjun Tu, Jiahao Lin, Qichao Zhang, Xiangyu Tian, Linjing Li, Xiangyuan Lan, and Dongbin Zhao.
\newblock Learning when to think: Shaping adaptive reasoning in r1-style models via multi-stage rl.
\newblock \emph{arXiv preprint arXiv:2505.10832}, 2025.

\bibitem[Wang et~al.(2024{\natexlab{a}})Wang, Pan, Shi, Lu, Zhan, and Li]{mathvision}
Ke~Wang, Junting Pan, Weikang Shi, Zimu Lu, Mingjie Zhan, and Hongsheng Li.
\newblock Measuring multimodal mathematical reasoning with math-vision dataset.
\newblock \emph{arXiv:2402.14804}, 2024{\natexlab{a}}.

\bibitem[Wang et~al.(2024{\natexlab{b}})Wang, Xia, He, Chen, Liu, Zhu, Liang, Wu, Liu, Malladi, et~al.]{wang2024charxiv}
Zirui Wang, Mengzhou Xia, Luxi He, Howard Chen, Yitao Liu, Richard Zhu, Kaiqu Liang, Xindi Wu, Haotian Liu, Sadhika Malladi, et~al.
\newblock Charxiv: Charting gaps in realistic chart understanding in multimodal llms.
\newblock \emph{Advances in Neural Information Processing Systems}, 37:\penalty0 113569--113697, 2024{\natexlab{b}}.

\bibitem[Xiao et~al.(2024)Xiao, Sun, Liu, and Wang]{xiao2024logicvista}
Yijia Xiao, Edward Sun, Tianyu Liu, and Wei Wang.
\newblock Logicvista: Multimodal llm logical reasoning benchmark in visual contexts.
\newblock \emph{arXiv preprint arXiv:2407.04973}, 2024.

\bibitem[Xu et~al.(2025)Xu, Wang, Wang, Chen, Zhou, Yang, Lu, Li, Wang, Zhu, et~al.]{xu2025visulogic}
Weiye Xu, Jiahao Wang, Weiyun Wang, Zhe Chen, Wengang Zhou, Aijun Yang, Lewei Lu, Houqiang Li, Xiaohua Wang, Xizhou Zhu, et~al.
\newblock Visulogic: A benchmark for evaluating visual reasoning in multi-modal large language models.
\newblock \emph{arXiv preprint arXiv:2504.15279}, 2025.

\bibitem[Yang et~al.(2025{\natexlab{a}})Yang, Li, Yang, Zhang, Hui, Zheng, Yu, Gao, Huang, Lv, et~al.]{yang2025qwen3}
An~Yang, Anfeng Li, Baosong Yang, Beichen Zhang, Binyuan Hui, Bo~Zheng, Bowen Yu, Chang Gao, Chengen Huang, Chenxu Lv, et~al.
\newblock Qwen3 technical report.
\newblock \emph{arXiv preprint arXiv:2505.09388}, 2025{\natexlab{a}}.

\bibitem[Yang et~al.(2025{\natexlab{b}})Yang, Wen, Liu, Chu, Song, Rao, Yi, Li, Zang, et~al.]{team2025kwai}
Biao Yang, Bin Wen, Changyi Liu, Chenglong Chu, Chengru Song, Chongling Rao, Chuan Yi, Da~Li, Dunju Zang, et~al.
\newblock Kwai keye-vl technical report.
\newblock \emph{arXiv preprint arXiv:2507.01949}, 2025{\natexlab{b}}.

\bibitem[Yu et~al.(2025)Yu, Zhang, Zhu, Yuan, Zuo, Yue, Dai, Fan, Liu, Liu, et~al.]{yu2025dapo}
Qiying Yu, Zheng Zhang, Ruofei Zhu, Yufeng Yuan, Xiaochen Zuo, Yu~Yue, Weinan Dai, Tiantian Fan, Gaohong Liu, Lingjun Liu, et~al.
\newblock Dapo: An open-source llm reinforcement learning system at scale.
\newblock \emph{arXiv preprint arXiv:2503.14476}, 2025.

\bibitem[Yu et~al.(2023)Yu, Yang, Li, Wang, Lin, Liu, Wang, and Wang]{yu2023mm}
Weihao Yu, Zhengyuan Yang, Linjie Li, Jianfeng Wang, Kevin Lin, Zicheng Liu, Xinchao Wang, and Lijuan Wang.
\newblock Mm-vet: Evaluating large multimodal models for integrated capabilities.
\newblock \emph{arXiv preprint arXiv:2308.02490}, 2023.

\bibitem[Yue et~al.(2023)Yue, Ni, Zhang, Zheng, Liu, Zhang, Stevens, Jiang, Ren, Sun, et~al.]{yue2023mmmu}
Xiang Yue, Yuansheng Ni, Kai Zhang, Tianyu Zheng, Ruoqi Liu, Ge~Zhang, Samuel Stevens, Dongfu Jiang, Weiming Ren, Yuxuan Sun, et~al.
\newblock Mmmu: A massive multi-discipline multimodal understanding and reasoning benchmark for expert agi.
\newblock \emph{arXiv:2311.16502}, 2023.

\bibitem[Yue et~al.(2024)Yue, Zheng, Ni, Wang, Zhang, Tong, Sun, Yin, Yu, Zhang, et~al.]{mmmupro}
Xiang Yue, Tianyu Zheng, Yuansheng Ni, Yubo Wang, Kai Zhang, Shengbang Tong, Yuxuan Sun, Ming Yin, Botao Yu, Ge~Zhang, et~al.
\newblock Mmmu-pro: A more robust multi-discipline multimodal understanding benchmark.
\newblock \emph{arXiv preprint arXiv:2409.02813}, 2024.

\bibitem[Yue et~al.(2025)Yue, Lin, Song, Wang, Ren, Gu, Li, Li, Zhao, Li, Bao, Tian, Zhang, Wang, Zhu, Cici, He, Ye, Shen, Zhang, Jiang, Zheng, Song, Luo, Yu, Wang, Tian, Tu, Yan, Huang, Wang, Xu, Song, Zhang, Yong, Zhang, Deng, Yang, Ma, Lv, Zhuang, Liu, Deng, Liu, Chen, Yu, Liu, Wang, Ma, Wang, Wang, Chen, Zhu, Zhou, Zhou, Fang, Shi, Dong, Xiao, Xu, Liu, Xu, Qu, Zhao, Lv, Wang, Zhang, Zhang, Zhang, Ma, Liu, Cai, and Xia]{coreteam2025mimovltechnicalreport}
Zihao Yue, Zhenru Lin, Yifan Song, Weikun Wang, Shuhuai Ren, Shuhao Gu, Shicheng Li, Peidian Li, Liang Zhao, Lei Li, Kainan Bao, Hao Tian, Hailin Zhang, Gang Wang, Dawei Zhu, Cici, Chenhong He, Bowen Ye, Bowen Shen, Zihan Zhang, Zihan Jiang, Zhixian Zheng, Zhichao Song, Zhenbo Luo, Yue Yu, Yudong Wang, Yuanyuan Tian, Yu~Tu, Yihan Yan, Yi~Huang, Xu~Wang, Xinzhe Xu, Xingchen Song, Xing Zhang, Xing Yong, Xin Zhang, Xiangwei Deng, Wenyu Yang, Wenhan Ma, Weiwei Lv, Weiji Zhuang, Wei Liu, Sirui Deng, Shuo Liu, Shimao Chen, Shihua Yu, Shaohui Liu, Shande Wang, Rui Ma, Qiantong Wang, Peng Wang, Nuo Chen, Menghang Zhu, Kangyang Zhou, Kang Zhou, Kai Fang, Jun Shi, Jinhao Dong, Jiebao Xiao, Jiaming Xu, Huaqiu Liu, Hongshen Xu, Heng Qu, Haochen Zhao, Hanglong Lv, Guoan Wang, Duo Zhang, Dong Zhang, Di~Zhang, Chong Ma, Chang Liu, Can Cai, and Bingquan Xia.
\newblock Mimo-vl technical report, 2025.
\newblock URL \url{https://arxiv.org/abs/2506.03569}.

\bibitem[Zhan et~al.(2025)Zhan, Deng, Tang, Xiang, Wu, Li, Zhu, Xu, Huang, Feng, et~al.]{zhan2025kat}
Zizheng Zhan, Ken Deng, Huaixi Tang, Wen Xiang, Kun Wu, Weihao Li, Wenqiang Zhu, Jingxuan Xu, Lecheng Huang, Zongxian Feng, et~al.
\newblock Kat-v1: Kwai-autothink technical report.
\newblock \emph{arXiv preprint arXiv:2507.08297}, 2025.

\bibitem[Zhang et~al.(2025)Zhang, Lin, Hou, Feng, and Li]{zhang2025adaptthink}
Jiajie Zhang, Nianyi Lin, Lei Hou, Ling Feng, and Juanzi Li.
\newblock Adaptthink: Reasoning models can learn when to think.
\newblock \emph{arXiv preprint arXiv:2505.13417}, 2025.

\bibitem[Zhang et~al.(2024)Zhang, Jiang, Zhang, Lin, Guo, Qiu, Zhou, Lu, Chang, Qiao, et~al.]{zhang2024mathverse}
Renrui Zhang, Dongzhi Jiang, Yichi Zhang, Haokun Lin, Ziyu Guo, Pengshuo Qiu, Aojun Zhou, Pan Lu, Kai-Wei Chang, Yu~Qiao, et~al.
\newblock Mathverse: Does your multi-modal llm truly see the diagrams in visual math problems?
\newblock In \emph{European Conference on Computer Vision}, pp.\  169--186. Springer, 2024.

\bibitem[Zhu et~al.(2025)Zhu, Wang, Chen, Liu, Ye, Gu, Tian, Duan, Su, Shao, et~al.]{zhu2025internvl3}
Jinguo Zhu, Weiyun Wang, Zhe Chen, Zhaoyang Liu, Shenglong Ye, Lixin Gu, Hao Tian, Yuchen Duan, Weijie Su, Jie Shao, et~al.
\newblock Internvl3: Exploring advanced training and test-time recipes for open-source multimodal models.
\newblock \emph{arXiv preprint arXiv:2504.10479}, 2025.

\bibitem[Zou et~al.(2024)Zou, Guo, Yang, Zhang, Hu, and Zhang]{zou2024dynamath}
Chengke Zou, Xingang Guo, Rui Yang, Junyu Zhang, Bin Hu, and Huan Zhang.
\newblock Dynamath: A dynamic visual benchmark for evaluating mathematical reasoning robustness of vision language models.
\newblock \emph{arXiv preprint arXiv:2411.00836}, 2024.

\end{thebibliography}
\bibliographystyle{iclr2025_conference}

\newpage
\appendix
\section*{Appendix}

\section{Contributions}

\paragraph{Project Lead}
Jie Jiang$^1$

\paragraph{Core Contributors}
Qi Yang$^{1,2}$, Bolin Ni$^{1}$

\paragraph{Supervisors}
Shiming Xiang$^{2}$, Han Hu$^{1}$, Houwen Peng$^{1}$

\setcounter{table}{0} 
\setcounter{figure}{0} 

\renewcommand{\thetable}{\Roman{table}}

\renewcommand{\thefigure}{\Roman{figure}} 
\section{Pre-training Stage}
\label{sec:pretraining_setting}

\subsection{\xvl{} Model Architecture}
Similar to other MLLMs~\citep{li2024llava, zhu2025internvl3, bai2025qwen2}, \xvl{} consists of three core components for foundational multimodal understanding. The visual encoder is initialized with SigLIP2-So400m~\citep{tschannen2025siglip} enhanced with the AnyRes strategy~\citep{liu2024improved}, enabling it to flexibly process images of any resolution. For language comprehension and generation, we utilize the Qwen3-4B Large Language Model~(LLM)~\citep{yang2025qwen3}, chosen for its strong reasoning capabilities and computational efficiency. To bridge these two modalities, a randomly initialized Multi-Layer Perceptron~(MLP) projector maps the visual features into the LLM's latent space, ensuring a coherent multimodal interaction.
Furthermore, to establish robust non-thinking multimodal understanding capabilities, we employ a similar three-stage pre-training strategy~\citep{li2024llava,chen2024expanding} with 1.~\textit{MLP warmup}, 2.~\textit{vision-language alignment}, and 3.~\textit{joint multimodal pre-training}, producing a base vision-language model with comprehensive non-thinking multimodal understanding. More details are provided in supplementary material.

\subsection{Pre-training Settings}
Our framework implements a three-stage pre-training paradigm, summarized in Table~\ref{tab:training_stages}.
\begin{table}[hbt]
\centering
\small
\resizebox{0.98\textwidth}{!}{
\begin{tabular}{@{}l|ccc|c@{}}
\toprule
\textbf{Stages} & \textbf{Stage 1} & \textbf{Stage 2} & \textbf{Stage 3} & \textbf{Stage 1} \\
\midrule
\multirow{2}{*}{Purpose} & 
MLP & 
Vision-Language & 
Joint Multimodal & 
Bi-Mode  \\
& Warmup & Alignment & Pre-training & Annealing \\ \midrule

Batch Size & 512 & 2048& 1024 & 1024 \\

Scheduler Type & Cosine & Cosine & Constant & Cosine\\

MLP Learning Rate & 1e-3 & 4e-5, 4e-6 & 4e-6 & 4e-6 \\

ViT Learning Rate & - & 4e-5, 4e-6 & 4e-6 & 4e-6 \\

LLM Learning Rate & 
- & 
- & %
4e-5 & 
4e-5,4e-6 \\

Packed Sequence Length & 
8192 & 
16384 & 
16384 & 
16384 \\

Trainable Components &  MLP & ViT, MLP & ViT, MLP, LLM & ViT, MLP, LLM \\

\bottomrule
\end{tabular}
}
\caption{Overview of \xvlbase{} pre-training and annealing stages, including MLP warmup, vision-language alignment, joint multimodal pre-training, and bi-mode annealing.}
\label{tab:training_stages}
\end{table}

\textbf{Stage 1: MLP warmup}
We begin by freezing the parameters of both the ViT and the LLM, while initializing a randomly-initialized two-layer MLP projection module. This projector is trained using image-caption pairs to establish initial cross-modal alignment. This stage enables stable gradient propagation in subsequent stages and mitigates instability caused by poorly aligned representations.

\textbf{Stage 2: Vision-language alignment}
In this stage, the ViT backbone is unfrozen while the LLM remains frozen, and training proceeds using interleaved multimodal data. The inclusion of diverse visual content in these batches systematically improves the visual encoder’s ability to handle different visual domains.

\textbf{Stage 3: Joint multimodal pre-training}
This stage enables full-parameter optimization across the entire architecture. We expand the training regimen to incorporate 145 billion tokens spanning diverse modalities and tasks, including OCR interpretation, visual grounding, mathematical reasoning, and structured data (tables/charts).
Additionally, we implement a non-thinking loss masking strategy. In this strategy, \texttt{<think> </think>} tags are appended before response generation, and their corresponding loss contributions are masked. This strategy effectively preserves Qwen3's~\citep{yang2025qwen3} specialized reasoning capabilities during joint multimodal pre-training.
\subsection{Non-Reasoning Data Distribution for Pre-training Stage}
\label{sec:non-reasoning_data}
To enhance \xvlbase{}'s capabilities, we employed diverse categories of data across distinct training stages. The pre-training stages primarily utilized non-reasoning data to improve multimodal understanding and visual perceptual abilities.

\textbf{Stage 1:} We trained the model using 808K image captioning samples sourced from the LAION datasets~\citep{liu2025ola}. The primary objective was to establish vision-language connections.

\textbf{Stage 2:} This stage focuses on refining the visual module and enhancing image understanding. We utilized a large corpus~(25 million items) comprising Visual OCR (31.8\%), Knowledge (33.7\%), Captioning (26.3\%), and Math (8.2\%) data.

\textbf{Stage 3:} To boost overall multimodal understanding, we introduced 13.3\% text data alongside higher proportions of quality Math and K-12 data (22.9\%). This stage included approximately 37 million items.

\section{Case Study}
\input{cases/case_5}
\input{cases/case_4}
\input{cases/case_2}
\input{cases/case_1}
\input{cases/case_3}
\input{cases/case_6}

\end{document}